\documentclass[10pt,twocolumn,letterpaper]{article}

\usepackage[pagenumbers]{cvpr} 

\usepackage{graphicx}
\usepackage{amsmath}
\usepackage{amssymb}
\usepackage{booktabs}
\usepackage[pagebackref,breaklinks,colorlinks]{hyperref}
\usepackage[capitalize]{cleveref}
\crefname{section}{Sec.}{Secs.}
\Crefname{section}{Section}{Sections}
\Crefname{table}{Table}{Tables}
\crefname{table}{Tab.}{Tabs.}

\usepackage{overpic}
\usepackage{enumitem} 
\usepackage{overpic} 
\usepackage{color}
\usepackage{mathtools} 
\usepackage{bbm}

\definecolor{turquoise}{cmyk}{0.65,0,0.1,0.3}
\definecolor{purple}{rgb}{0.65,0,0.65}
\definecolor{dark_green}{rgb}{0, 0.5, 0}
\definecolor{orange}{rgb}{0.8, 0.6, 0.2}
\definecolor{red}{rgb}{0.8, 0.2, 0.2}
\definecolor{darkred}{rgb}{0.6, 0.1, 0.05}
\definecolor{blueish}{rgb}{0.0, 0.3, .6}
\definecolor{light_gray}{rgb}{0.7, 0.7, .7}
\definecolor{pink}{rgb}{1, 0, 1}
\definecolor{greyblue}{rgb}{0.25, 0.25, 1}

\DeclareMathOperator*{\argmin}{arg\,min}

\newcommand{\loss}[1]{\mathcal{L}_\text{#1}}

\newcommand{\real}{\mathbb{R}}

\newcommand{\Figure}[1]{Figure~\ref{fig:#1}}

\usepackage{lipsum}

\usepackage{blindtext}

\newcommand{\things}{\textit{things}\xspace}
\newcommand{\thing}{\textit{thing}\xspace}
\newcommand{\stuff}{\textit{stuff}\xspace}

\newcommand{\chk}{$\checkmark$}

\usepackage{algorithm}
\usepackage{algpseudocode}

\def\cvprPaperID{2746} 
\def\confName{CVPR}
\def\confYear{2022}

\begin{document}

\title{Panoptic Neural Fields: A Semantic Object-Aware Neural Scene Representation}

\author{Abhijit Kundu~\textsuperscript{1} \quad
Kyle Genova~\textsuperscript{1} \quad
Xiaoqi Yin~\textsuperscript{1} \quad
Alireza Fathi~\textsuperscript{1} \quad
Caroline Pantofaru~\textsuperscript{1} \quad \\
Leonidas Guibas~\textsuperscript{1,4} \quad
Andrea Tagliasacchi~\textsuperscript{1,3} \quad 
Frank Dellaert~\textsuperscript{1,2} \quad
Thomas Funkhouser~\textsuperscript{1}
\\[1em]
\textsuperscript{1}{Google Research}
\quad
\textsuperscript{2}{Georgia Tech}
\quad
\textsuperscript{3}{Simon Fraser University}
\quad
\textsuperscript{4}{Stanford University}
\vspace*{-5mm}
}

\twocolumn[{%
\renewcommand\twocolumn[1][]{#1}%
\maketitle
\begin{center}
\centering
\captionsetup{type=figure}
\includegraphics[width=0.98\linewidth]{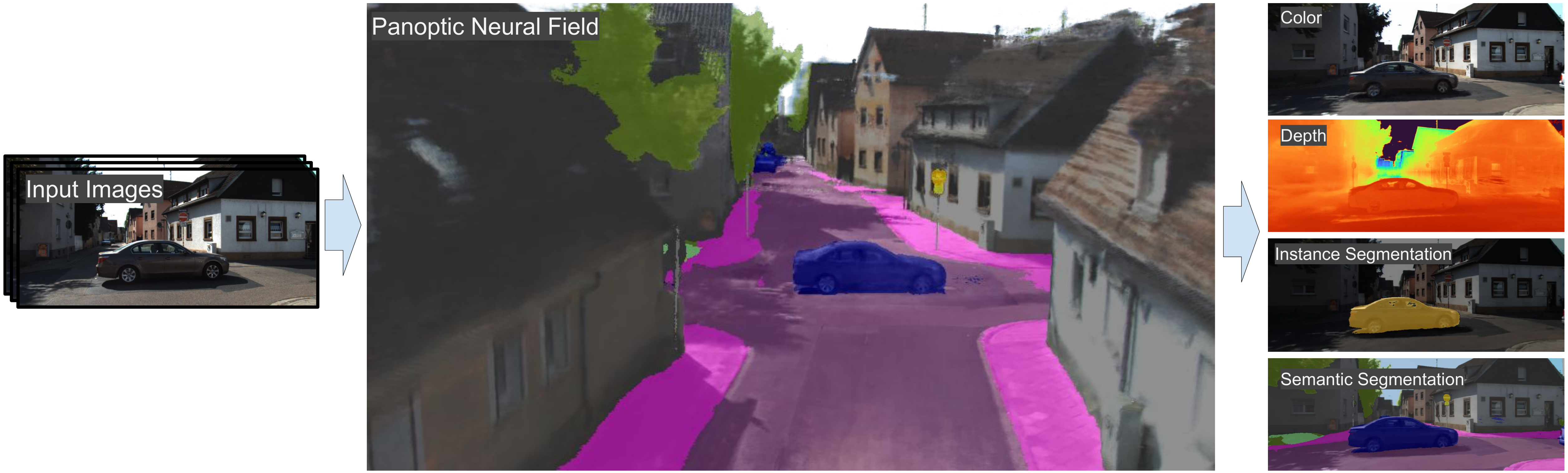}
\captionof{figure}{
\textit{Panoptic Neural Fields (PNF)} is an object-aware neural scene representation that decomposes a dynamic 3D scene into a set of objects (\things) and background (\stuff), each represented by a separate MLP based neural function. Our model predicts a panoptic-radiance field that represents the color, density, instance, and category label of any 3D point at any given time. The model is trained from RGB images alone and can describe dynamic challenging 3D scenes as shown above.}

%
\label{fig:teaser}
\end{center}%
}]
\maketitle
\begin{abstract}
We present Panoptic Neural Fields (PNF), an object-aware neural scene representation that decomposes a scene into a set of objects (things) and background (stuff).  Each object is represented by an oriented 3D bounding box and a multi-layer perceptron (MLP) that takes position, direction, and time and outputs density and radiance.  The background stuff is represented by a similar MLP that additionally outputs semantic labels. Each object MLPs are instance-specific and thus can be smaller and faster than previous object-aware approaches, while still leveraging category-specific priors incorporated via meta-learned initialization. Our model builds a panoptic radiance field representation of any scene from just color images. We use off-the-shelf algorithms to predict camera poses, object tracks, and 2D image semantic segmentations. Then we jointly optimize the MLP weights and bounding box parameters using analysis-by-synthesis with self-supervision from color images and pseudo-supervision from predicted semantic segmentations. During experiments with real-world dynamic scenes, we find that our model can be used effectively for several tasks like novel view synthesis, 2D panoptic segmentation, 3D scene editing, and multiview depth prediction.
\end{abstract}
\vspace{-0.2in}
\section{Introduction}
\label{sec:introduction}

The ability to understand the content within an image is an essential task in computer vision, and over time we have witnessed a rapid increase in task complexity. Over a short period of time, we have progressed from the task of identifying the overall presence of objects within an image~(i.e.~classification~\cite{krizhevsky2012imagenet} and object detection~\cite{He17iccv_MaskRCNN,girshick2015fast}), to fine grained pixel-by-pixel classification~(i.e.~semantic segmentation~\cite{long2015fully,ronneberger2015u}), and to the ability to differentiate between object instances of the same class (\ie panoptic segmentation~\cite{kirillov2019panoptic,Cheng20cvpr_Panoptic}).

However image level representations described above have limited applications. Instead we are interested in full 3D scene understanding which is  important for autonomous driving \cite{janai2020computer}, semantic mapping \cite{xia2020survey}, and many other applications involving navigation or operation in the physical world \cite{crespo2020semantic}. Given a sequence of RGB images, our goal is to infer: 1) a 3D reconstruction of the observed geometry, 2) a radiance field of the scene, 3) a decomposition of the scene into potentially dynamic things (e.g., cars) and background stuff (e.g., grass), 4) a category and instance label for every 3D point, as illustrated in \Figure{teaser}.

In recent years, neural 3D scene representations like NeRF~\cite{Mildenhall20eccv_nerf} have made significant advancements~\cite{tewari2021advances,xie2021neural}. NeRF represents a scene using a multi-layer perceptron (MLP) that maps positions and directions to densities and radiances which can then be used to synthesize an image from a novel view. However NeRF lacks semantic understanding and is also not object aware. In this work we explore neural scene representations for semantic 3D scene understanding tasks beyond the usual view synthesis task.

Some recent work augments NeRF to infer semantics \cite{Zhi21iccv_SemanticNeRF}, adding an extra head to predict semantic logits for any 3D position along with the usual density/color.
Other recent work decomposes a scene into a set of NeRFs associated with foreground objects separated from the background \cite{Guo20arxiv_OSF, Yang21iccv_ObjectNeRF, nsg21cvpr}.  However, these systems have several limitations in the context of our goals: 1) they do not produce panoptic segmentations, 2) they learn from scratch for every scene; and 3) they share MLPs for multiple objects, which limits their ability to reproduce specific instances.

We address these issues in our proposed Panoptic Neural Fields (PNF), an object-aware neural scene representation that explicitly decomposes a scene into a set of objects (\textit{things}) and amorphous \textit{stuff} background. Each object instance is represented by a \emph{separate} MLP to evaluate the radiance field within the local domain of a potentially moving and semantically labeled 3D bounding box. The semantic-radiance field of the \textit{stuff} background is also represented by a MLP which includes an additional semantic head. Together the \textit{stuff} and \textit{things} MLPs jointly define a panoptic-radiance field that describes the density, color, category, and instance label of any 3D point over time.

Our object aware representation makes it possible to describe scenes with multiple moving objects and also paves the way to incorporate constraints that objects of the same category have similar shape and appearance. Previous object-aware frameworks \cite{nsg21cvpr, Yang21iccv_ObjectNeRF} used a shared MLP with instance-specific latent codes to incorporate this prior. In our model, each object instance is represented by a \emph{separate} MLP that is initialized with a category-specific prior using meta-learning.  The separation of learning of object category priors via meta-learning makes it possible to represent instance-specific details with smaller MLPs, which speeds inference in scenes with many objects.


Given a collection of images captured from a scene, we employ off-the-shelf algorithms to predict camera parameters~\cite{MurArtal15tro_OrbSLAM} and 2D semantic segmentations~\cite{Cheng20cvpr_Panoptic} for all images, plus a set of 3D object detections with 3D oriented bounding boxes and category labels~\cite{Park21iccv_DD3D}. We initialize the weights of the MLPs for our panoptic neural field model either with object category-specific meta-learned initialization or simple biased initialization of density activation layers.  We then jointly optimize the bounding box and MLP parameters to minimize \textit{analysis-by-synthesis} style losses that measure differences in color and semantic images synthesized with volumetric rendering (as in NeRF \cite{Mildenhall20eccv_nerf}). Thus, our approach provides an unified framework for optimizing 3D shape, appearance, semantics, and object poses all from a set of color images.


We evaluate our method on several scene understanding and synthesis tasks using experiments on the KITTI\cite{Geiger12cvpr_KITTI} and KITTI-360~\cite{kitti360} dataset, including 3D panoptic reconstruction, and scene editing.  The output panoptic-radiance field can also be used to synthesize 2D image-level outputs like semantic segmentation, panoptic segmentation, depth images, and colored images of both observed and novel views.
We demonstrate the utility of the proposed method for these scene understanding tasks, as well as for novel-view synthesis method with movable scene components.



Our contributions can be summarized as follows:
\begin{itemize}
\setlength\itemsep{-.3em}
\item We propose, to the best of our knowledge, the first method that can derive a panoptic-radiance field of complex dynamic 3D scenes from images alone.
\item Our single unified model achieves state-of-the-art quality across multiple tasks and benchmarks on KITTI and KITTI-360 datasets.
\item We incorporate object shape and appearance priors via category-specific meta-learned initialization. This allows our object MLPs to be much smaller and faster than previous object-aware representations.
\item We jointly optimize all (\textit{stuff} and \textit{things}) neural fields and object poses, allowing our method to cope with noisy object poses and image segmentations.
\end{itemize}

\section{Related Work}
\label{sec:related_works}

The most relevant related work is summarized in Table~\ref{tab:related_work} and can be broadly divided into three categories: (1) Learning based single image 3D semantic and/or instance segmentation, (2) Multi-view 3D reconstruction and segmentation methods, and (3) Neural fields.

\textbf{Single image reconstruction and segmentation.}
3D-RCNN~\cite{3drcnn} and Mesh-RCNN~\cite{Gkioxari19iccv_MeshRCNN} takes as input a single RGB image and predicts 3D mesh and pose of object instances in the image. Total3DUnderstanding~\cite{Nie20cvpr_Total3D} combines layout estimation, 3D object detection, and object mesh generation. More recently~\cite{Dahnert21neurips_Panoptic3D} showed 3D panoptic reconstruction and segmentation from a single RGB image.

\definecolor{dark-green}{rgb}{0.0, 0.5, 0.0}
\newcommand{\chg}{\textcolor{dark-green}{\chk}}

\begin{table}[t]
\centering
\small
\begin{tabular}{p{3cm}p{4mm}p{4mm}p{4mm}p{4mm}p{4mm}p{4mm}}
\hline
 Paper                                                     & Sem   & Obj   & Pan   & Dyn   & Opt   &  Syn  \\
\hline
 MeshRCNN~\cite{Gkioxari19iccv_MeshRCNN}                   &       & \chk  &       &       &       &       \\
 Total3D~\cite{Nie20cvpr_Total3D}                          & \chk  & \chk  &       &       &       &       \\
 \hline
 Atlas~\cite{Murez20eccv_atlas}                            & \chk  &       &       &       &       &       \\
SLAM++\cite{SalasMoreno13icra_slam}                      &       & \chk. &       &       &       &       \\
PanopticFusion~\cite{Narita19iros_Panopticfusion}          & \chk  &       & \chk  &       &       &       \\
 Kimera~\cite{Rosinol20icra_Kimera}                        & \chk  &       &       &       &       &       \\
 DynSceneGraphs~\cite{Rosinol20rss_dynamicSceneGraphs} & \chk  & \chk  & \chk  & \chk  & \chk  &       \\
\hline
 SemanticNerF~\cite{Zhi21iccv_SemanticNeRF}                & \chk  &       &       &       &       & \chk \\
 NSG~\cite{nsg21cvpr}                                  &       & \chk  &       & \chk  &       & \chk \\
 ObjectNeRF~\cite{Yang21iccv_ObjectNeRF}                   &       & \chk  &       &       &       & \chk \\
\hline
 PNF (Ours)                                       & \chg  & \chg  & \chg  & \chg  & \chg  & \chg \\
\hline
\end{tabular}
\caption{Comparison to properties of related work.   The check marks indicate which prior methods have the following capabilities:   
``Sem'' = performs semantic segmentation;
``Obj'' = performs object decomposition;
``Pan'' = performs panoptic segmentation;
``Dyn'' = handles dynamic objects; and
``Opt'' = optimizes object bbox parameters.
``Syn'' = allows for novel view synthesis.
}
\label{tab:related_work}
\end{table}

\textbf{Multi-view reconstruction and segmentation:} Incorporating semantics into SLAM and SfM systems has a long history~\cite{hane2013joint,kundu2014joint,blaha2016large,SalasMoreno13icra_slam}.
More recently, PanopticFusion~\cite{Narita19iros_Panopticfusion} is an incremental, online mapping approach that fuses a sequence of RGB-D images into a consistent panoptic segmentation.
ATLAS \cite{Murez20eccv_atlas} reconstructs and labels the 3D geometry from \textit{multiple} posed RGB images. However, ATLAS produces only semantic segmentations (without instances). Both PanopticFusion and ATLAS works only for static scenes, requires 3D supervision, and relies upon convolutions on a discrete voxel grid, which limits its resolution. Kimera~\cite{Rosinol20icra_Kimera} takes a stereo sequence and does online reconstruction, meshing, and semantic labeling of the mesh using \textit{ ground-truth labels}, as a proxy for any 2D segmentation method.
Dynamic Scene Graphs~\cite{Rosinol20rss_dynamicSceneGraphs} expands on that by inferring object instances, even dynamic ones in case of people.
Both methods, though representing impressive systems, were only demonstrated in simulation and rely on ground truth semantic labels.

\textbf{Neural radiance fields (NeRFs).}
This work builds upon NeRF \cite{Mildenhall20eccv_nerf}, which represents a scene using a multi-layer perceptron (MLP) that maps positions and directions to densities and radiances.
From that representation, novel views can be synthesized using volumetric rendering and compared to input views in a self-supervised optimization procedure to infer MLP weights for an observed scene.
However, NeRF only works for static scenes and trains for hours (from scratch) for every set of input views.

\textbf{NeRFs with semantics.} 
Recent work has considered using neural representations to infer semantics \cite{Zhi21iccv_SemanticNeRF}.  In particular, SemanticNeRF \cite{Zhi21iccv_SemanticNeRF} adds an extra head to NeRF to predict semantic labels for any 3D position along with the usual density and color. Concurrent to our work ~\cite{fu2022panoptic} also demonstrated neural panoptic fusion from multiple views. However both of these work are not object aware and cannot handle dynamic scenes.

\textbf{NeRFs with dynamics} disentangle a scene into a canonical volume and its time-varying deformation, represented by a second MLP.
This approach has been applied for deforming faces \cite{Gao20arxiv_pNeRF, Gafni21cvpr_DNRF, Raj21cvpr_ANR}, moving human bodies \cite{Weng20arxiv_vid2actor, Peng21cvpr_NeuralBody, Su21arxiv_A-NeRF}, and objects \cite{Du21iccv_NeRFlow, Li21cvpr_nsff, Park21iccv_nerfies, Pumarola21cvpr_D_NeRF,Xian20cvpr_stnif}.
In contrast, we consider dynamic scenes that contain many moving objects.

\textbf{NeRFs with object decompositions}~\cite{Guo20arxiv_OSF, Yang21iccv_ObjectNeRF, nsg21cvpr} decompose a scene into a set of NeRFs associated with foreground objects separated from the background.
ObjectNeRF~\cite{Yang21iccv_ObjectNeRF} 
uses the object branch to render rays with masked areas for foreground objects conditioned on a latent code.
Similarly, Neural Scene Graphs (NSGs)~\cite{nsg21cvpr} 
uses a separate conditional NeRF for each object category, and a multiplane neural representation for the background.
However, these systems have several limitations in the context of our goals: 1) they do not produce panoptic segmentations, 2) they learn from scratch for every scene; and 3) they share NeRFs for multiple objects (which limits the ability to reproduce specific discrete instances).

\textbf{Conditional NeRFs} infer latent codes as pioneered in GRAF~\cite{Schwarz20neurips_graf}, piGAN~\cite{Chan21cvpr_piGAN}, and PixelNeRF~\cite{Yu21cvpr_pixelNeRF}, as well as the recent CodeNeRF~\cite{Jang21iccv_CodeNeRF}, which also optimizes over object poses. All these works incorporate category-specific priors by sharing MLP weights across object instances, combined with instance specific codes. We instead use instance-specific MLPs for representing each object, which allows each MLP to be smaller, resulting in faster inference speed on scenes with multiple objects. Object appearance and shape priors are incorporated via category-specific meta-learned initialization~\cite{Tancik21cvpr_meta,finn2017maml,nichol2018reptile,jiang2019improving} of the MLP weights.

\newcommand{\image}{\mathcal{C}}
\newcommand{\C}{C} 
\newcommand{\gt}{\text{gt}} 
\newcommand{\params}{\boldsymbol{\theta}}
\newcommand{\near}{{t_n}}
\newcommand{\far}{{t_f}}
\newcommand{\radiance}{\mathbf{c}}
\newcommand{\sky}{\text{sky}}
\newcommand{\ray}{\mathbf{r}}
\newcommand{\origin}{\mathbf{o}}
\newcommand{\dir}{\mathbf{d}}
\newcommand{\gdir}{\gamma(\dir)}
\newcommand{\density}{\boldsymbol{\sigma}}
\newcommand{\feature}{\mathbf{z}}
\newcommand{\definedAs}{\coloneqq}
\newcommand{\track}{\mathbf{T}}
\newcommand{\tracks}{\track_{\iThing,\iImage}}
\newcommand{\iThing}{k}
\newcommand{\nThings}{K}
\newcommand{\instance}{\mathbf{i}}
\newcommand{\instanceMap}{\mathcal{I}}
\newcommand{\iImage}{i}
\newcommand{\nImages}{I}
\newcommand{\given}{\,|\,}
\newcommand{\depth}{\mathcal{D}}
\newcommand{\point}{\mathbf{x}}
\newcommand{\gpoint}{\gamma(\point)}
\newcommand{\timestamp}{\mathrm{t}}
\newcommand{\semantic}{\mathbf{s}}
\newcommand{\nClasses}{C}
\renewcommand{\S}{\mathcal{S}}
\newcommand{\rotation}{\mathbf{R}}
\newcommand{\translation}{\mathbf{t}}
\newcommand{\size}{\mathbf{s}}
\newcommand{\scale}{\mathbf{s}}

\begin{figure*}[t]
\centering
\includegraphics[width=0.935\linewidth]{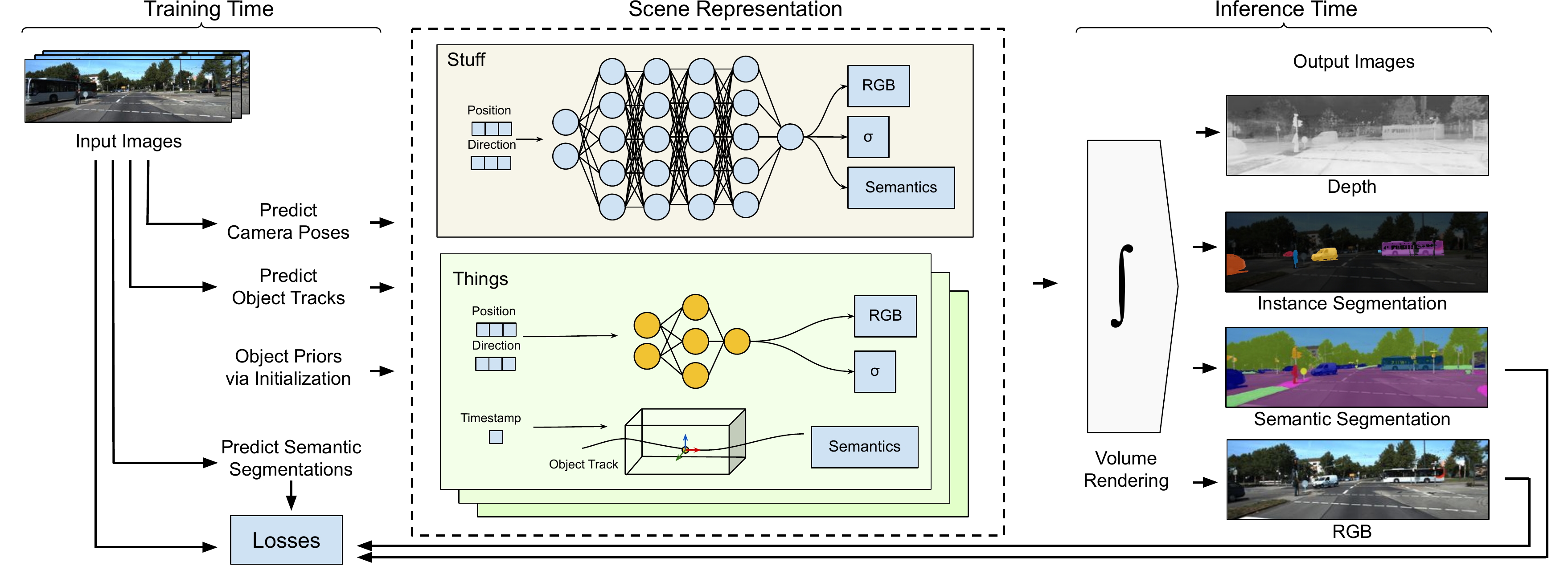}
\vspace{-0.8em}
\caption{
    Overview of \textit{Panoptic Neural Feld (PNF)} representation being learned from input color images. The background \textit{stuff} is represented by an MLP that produces RGB, density, and semantic logits, while each object is represented by a dynamic track and a smaller individual MLP. Once trained the representation can be used for several tasks by simple volume rendering.
}
\label{fig:system_overview}
\end{figure*}
\section{Method}
\label{sec:method}

This section introduces the panoptic neural field representation and our computational pipeline (Figure~\ref{fig:system_overview}). In ~\cref{sec:scene_rep} we describe the representation itself, which stores a panoptic-radiance field that can be used to query the color, density, semantic and instance labels at any 3D point at any time.  In~\cref{sec:rendering}, we describe how this panoptic-radiance field can be rendered using NeRF-style volume rendering by \textit{over} compositing along sampling of points along rays. In~\cref{sec:model_train} we explain how the model is trained in analysis-by-synthesis style by comparing the rendered color and semantic segmentation with observed 2D color and predicted 2D semantic labels.

\begin{figure}[t]
\centering
\includegraphics[width=1.0\linewidth]{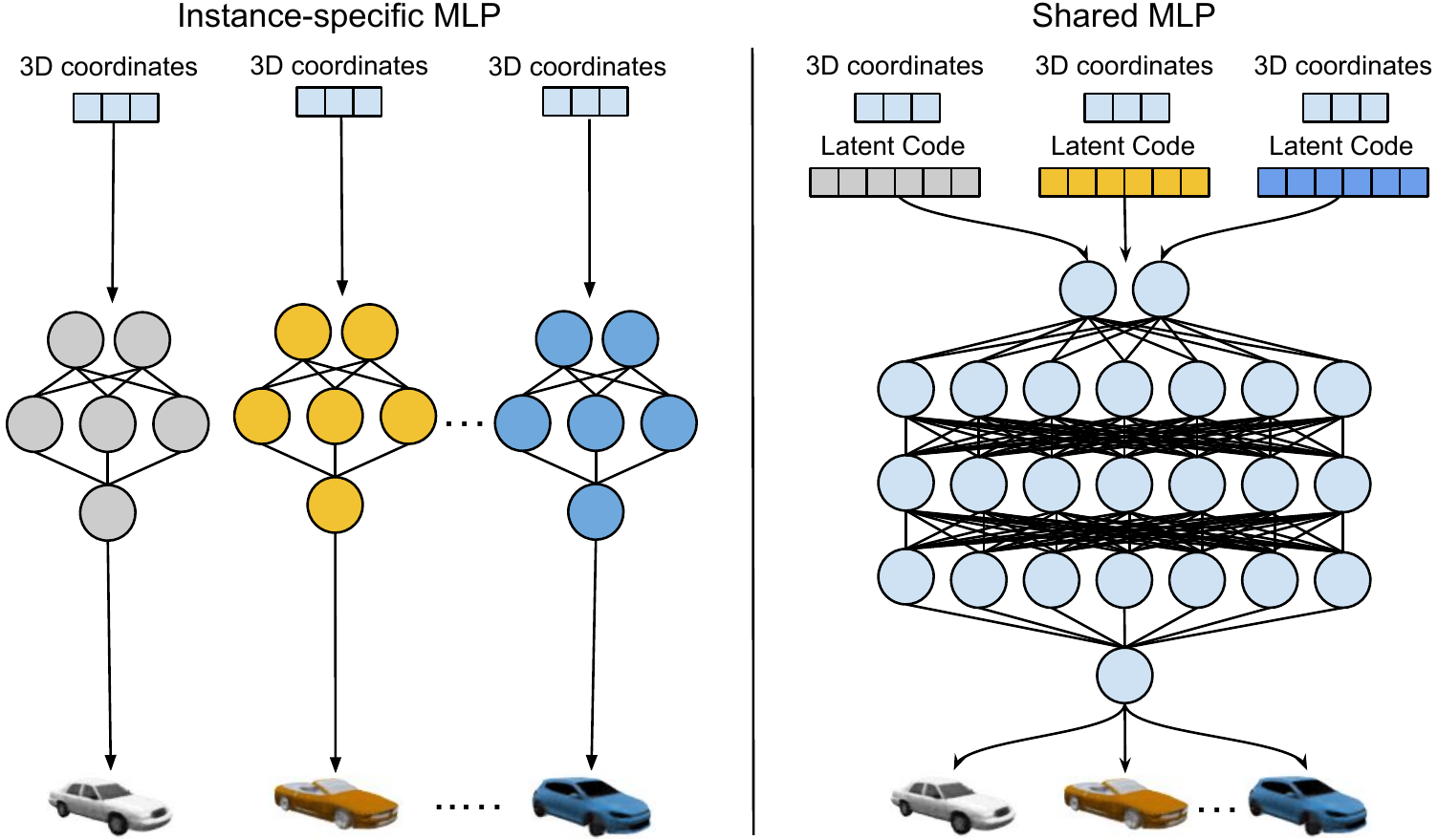}
\caption{Our framework uses instance-specific fully weight encoded functions to represent each object (left), in comparison to a traditional approach of using one shared MLP with instance-specific latent codes (right). Instance-specific MLPs can be smaller, since they just need to have enough capacity to express a single object instance, as compared to a shared MLP. On scenes with multiple objects our approach can be significantly faster.
}
\label{fig:instance_specific_neural_function}
\end{figure}

A key difference between our framework and previous object-aware frameworks \cite{nsg21cvpr, Yang21iccv_ObjectNeRF} is how we train and represent \things. As illustrated in ~\cref{fig:instance_specific_neural_function}, our framework uses instance-specific fully weight encoded functions to represent each object, in comparison to the traditional approach of using a shared MLP with instance-specific latent codes. This design choice is driven by several factors. First, since the MLP only needs to represent a single object instance, we can have a smaller MLP compared to shared MLPs, resulting in faster inference speed on scenes with multiple objects. Second, this allows the object MLP to use its full capacity to describe and overfit to a specific novel object instance, which may not be possible~\cite{davies2020overfitsdf} with latent encoding. Third, it is simpler and does not require any change to the core NeRF model architecture. Object-level priors can also be incorporated to our instance-specific models using meta-learning based initialization (See \cref{sec:model_init}).

\subsection{Scene Representation}
\label{sec:scene_rep}

The core of our framework is the panoptic-radiance field representation. This representation accepts input queries consisting of a point position $ \point \in \mathbb{R}^3$, view direction $\dir \in \mathbb{R}^3$, and time $\timestamp \in \mathbb{R}$. The outputs of a query are color, density, a semantic label, and an instance label. This field is the composition of multiple distinct neural functions. There are separate fields for each 3D object (\things), and another, larger field for the background (\stuff). The field associated with one object is defined inside a mobile 3D oriented bounding box. The background is represented by another neural function defined inside a larger scene bounding box. It encodes density, appearance, and semantic labels.

\textbf{Things:}
Foreground objects in our representation are represented by a neural function inside a dynamic bounding box. To instantiate the set of object tracks in a scene, we first run an RGB-only 3D object detector~\cite{Park21iccv_DD3D} and tracker~\cite{pc3t}. This provides a bounding box track $T_{k}$ and semantic class for each recognized object instance $k$. The track is parameterized by a sequence of transformation matrices, one at each of at a set of discrete timestamps. For each timestamp, we create a rotation matrix $\rotation \in \real^{3 \times 3}$ and a translation vector $\translation \in \real^3$. There is also a box extent $\size \in \real^3$ along each axis that is time-invariant.  To determine the coordinate frame of an object at an arbitrary real-valued timestamp, we interpolate the discrete track steps. 

For each object instance, we instantiate a separate time-invariant MLP with the standard NeRF architecture~\cite{Mildenhall20eccv_nerf}. Its weights are initialized using the techniques described in Section~\ref{sec:model_init}.  To query this MLP, positions and directions are transformed from the world frame to the bounding box frame 
defined by the track at the current timestamp.
We optimize all parameters of the MLP and object track $T_{k}$ jointly. Optimizing object track parameters is important as initial boxes (even GT boxes) may be noisy. In order to optimize rotation, we orthogonalize $\rotation$ after each gradient descent step using SVD, which projects it back onto $\mathrm{SO}(3)$.


\textbf{Stuff:}
We represent the static background \stuff with a single neural function. In addition to predicting density and color at every 3D point, the \stuff function also learns a semantic label per point. We again use an MLP to represent the learned function. The architecture is similar to NeRF, but with an additional head for semantic logits. This head is direction-invariant to encode the inductive bias that 3D points have multi-view consistent semantic labels. Note that unlike the MLPs for objects, which are bounded, the \stuff MLP must handle the unbounded nature of real world scenes. Therefore for large scenes, we follow~\cite{zhang2020nerf++} and use a separate foreground and background \stuff MLPs.

\textbf{Panoptic-Radiance Field:}
The final panoptic-radiance field at a 3D point is computed from aggregating the contributions of the individual \textit{thing} and \stuff MLPs. For any given output channel (color, density, etc.), our function takes the sum of all contributions from any bounding box hits, defaulting to the \stuff output if there is no intersection. For the color field $\radiance$, this is: 
\begin{align}
\radiance(\point \given \params) &= \mathbbm{1}_S(\point)\radiance_{s}(\point \given \params) + \sum_\iThing \radiance_\iThing (\track^{-1}_\iThing \point \given \params)
\end{align}
 where $\mathbbm{1}_S$ is 1 if and only if the point intersects no bounding boxes, $\radiance_s$ is the \stuff color field, and $\params$ denote the MLP weights for \stuff and \thing MLPs. For other fields, we simply substitute the radiance $\radiance$ with the density, semantic, or instance function. Object boxes contribute a one-hot semantic logit vector for their class, which handles the merging of \stuff and \textit{thing} semantics.
 Similarly, the instance label function is a vector of length $K$, with one dimension per detected object $k$. Objects contribute a one-hot vector for their instance while the \stuff function's instance output is always zero.

\subsection{Rendering Panoptic-Radiance Fields}
\label{sec:rendering}

Given the complete panoptic-radiance field representation, a 2D image can be synthesized with volume rendering. This process is described in more detail in NeRF~\cite{Mildenhall20eccv_nerf}. Our image synthesis approach is the similar to NeRF, with the addition of support for extra output channels and dynamic boxes. To render a single ray $\ray = \mathbf{o} + t \mathbf{d}$ we uniformly sample $N=1024$ points $\point$ (with jitter) along the ray and alpha-composite the result of the output channel $\C$ we wish to render (RGB, depth, semantic, instance):
\begin{align}
\C(\ray \given \params) \approx \sum_{i=1}^{N} w(t_i)f(\ray(t_i) \given \params).
\label{eq:render}
\end{align}
Above, $w(t)$ is the final weight associated with each sample, determined by \textit{over} compositing the opacity values of each sample along the ray. The function $f$ returns the representation value for the channel in question at the query point. For semantics, this is logits, while for instance it is a one-hot encoding of the object instance identifier $k$.

\subsection{Model Losses and training}
\label{sec:model_train}

We jointly optimize all network parameters $\theta$ and object tracks $\track$ to reproduce the observed RGB images and predicted 2D semantic images:
\begin{align}
\argmin_{\params,\tracks} \:\:
&\loss{rgb} (\params,\tracks)
+ 
\loss{sem} (\params,\tracks)
\label{eq:optim}
\end{align}
At each gradient descent step, we randomly sample mini-batches of rays. Our RGB loss is the mean squared error between the synthesized and ground truth color, summed over sampled rays as in NeRF~\cite{Mildenhall20eccv_nerf}. Our semantic loss is applied at the same pixel locations, and compares the synthesized semantics with the input 2D semantic segmentation prediction~\cite{Cheng20cvpr_Panoptic}. For this loss, we apply a per-pixel softmax-cross entropy function rather than mean squared error.






\begin{figure}[t]
\centering
\includegraphics[width=1.0\linewidth]{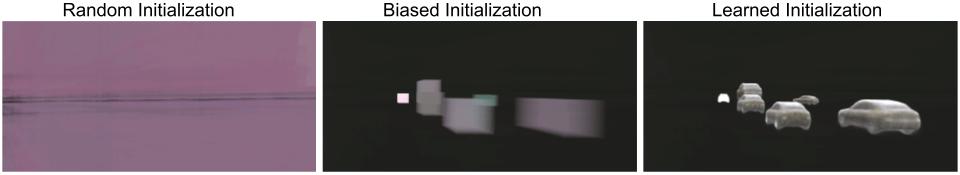}
\caption{This figure illustrates different initialization schemes of instance-specific MLP weights for ``things'' in the car category.}
\label{fig:init_illustration}
\end{figure}
\begin{figure}[t]
\centering
  \includegraphics[width=1.0\linewidth]{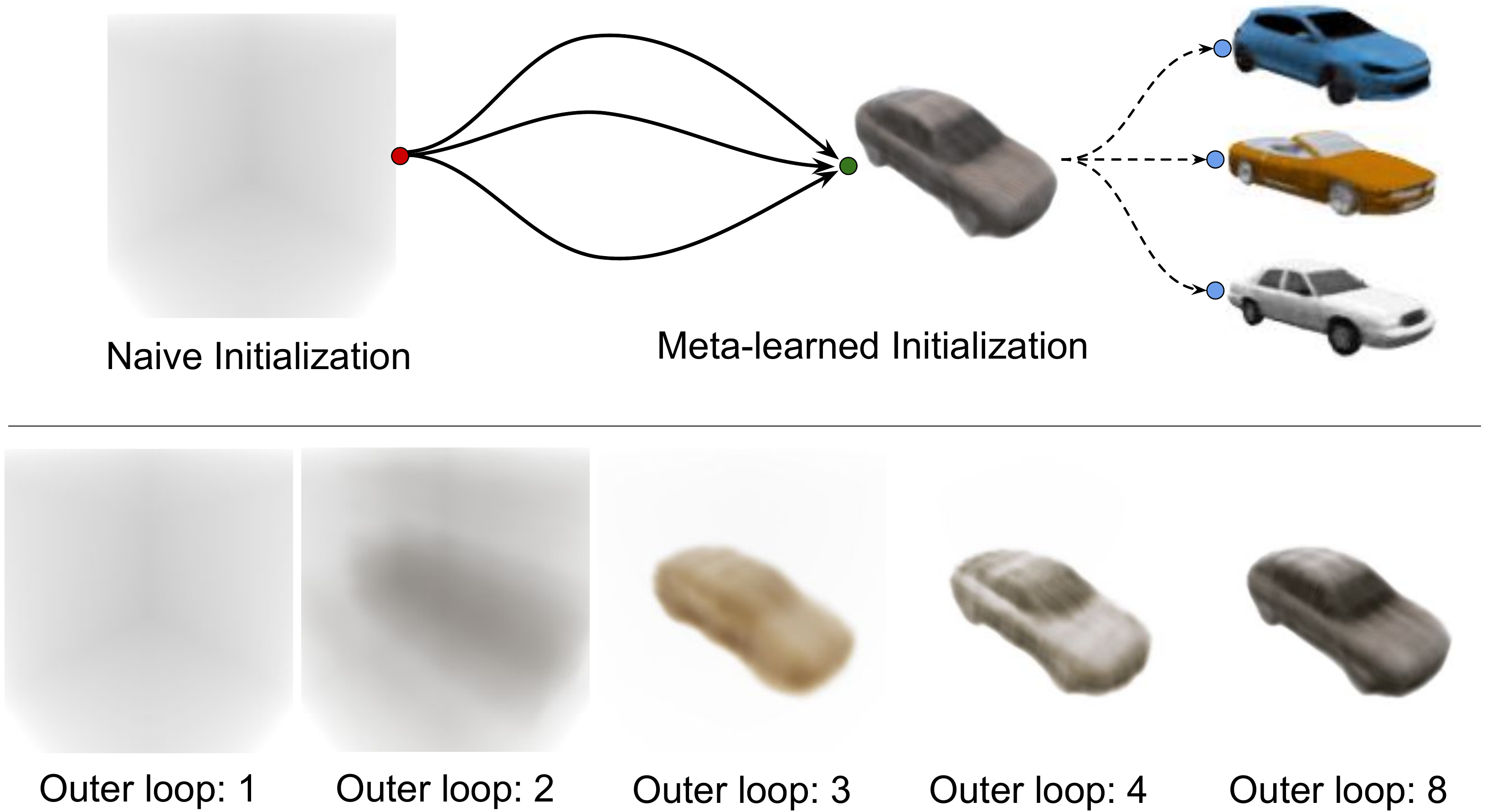}
\caption{Category specific learned initialization with federated averaging. \textbf{Upper:} Rather than random initialization of MLP weights, we use meta-learning implemented using FedAvg~\cite{mcmahan2017fedavg,jiang2019improving} to find a good category specific initialization. This allows improves generalization and convergence behavior. \textbf{Lower:} We visualize the learned initialization model for the car class with increasing outer loops of the federated averaging algorithm.}
\label{fig:fed_avg}
\end{figure}

\subsection{Incorporating priors via initialization}
\label{sec:model_init}
One of the core benefits of an object aware approach, is the ability to incorporate inductive bias that objects instances within same category, often have similar 3D shape and appearance. One possible way to incorporate such priors is to have shared MLP weights across all object instances, combined with some instance specific codes. Our framework instead uses separate MLPs for representing each object instance. As illustrated in ~\cref{fig:instance_specific_neural_function}, this allows each MLP to be smaller as it only needs to represent a single object instance, resulting in faster inference speed on scenes with multiple objects. Object shape and appearance priors are instead incorporated via initialization of the MLP weights of the neural functions. We present two approaches (see \cref{fig:init_illustration}) of initializing our model, one based on category specific meta-learning and another based on simple bias initialization of the activation function of the MLPs.


%
\begin{figure*}
  \centering
  \includegraphics[width=\linewidth]{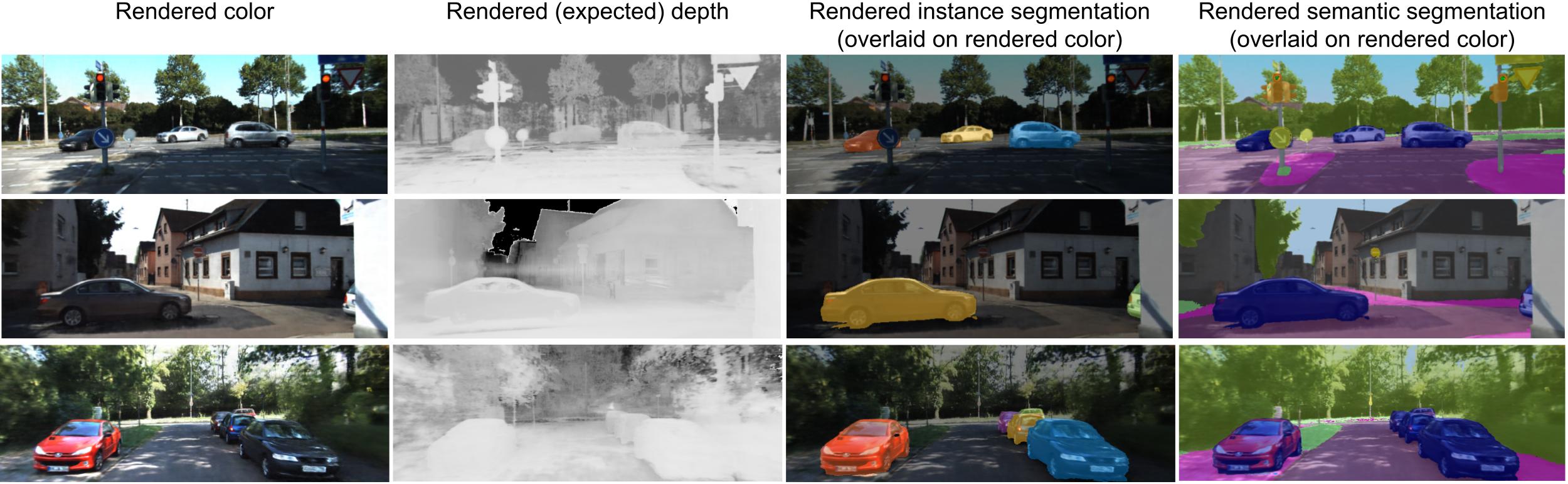}
  \vspace*{-6mm}
   \caption{Images of color, depth, instance segmentation, and semantic segmentation rendered on dynamic KITTI scenes from our model trained \textbf{only} from RGB images.}
   \label{fig:renders_2d_predicted_boxes}
\end{figure*}
\begin{figure*}
  \centering
  \includegraphics[width=\linewidth]{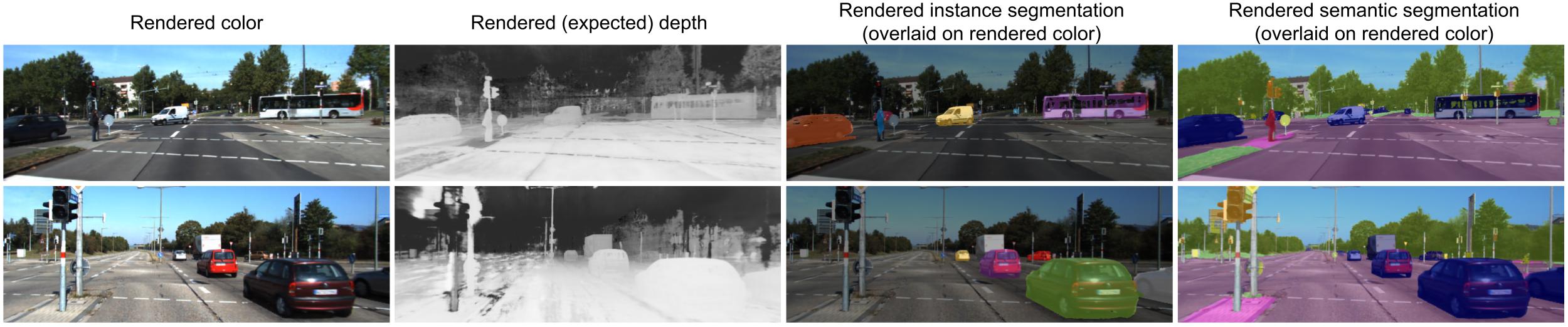}
  \vspace*{-6mm}
   \caption{Rendered color, depth, instance segmentation, and semantic segmentation images of our models representing dynamic KITTI scenes with various object categories. These results used KITTI \textbf{provided} object tracks. Note that our instance specific object MLPs can also reconstruct novel object categories  like \textit{truck} and ~\textit{bus}.}
   \label{fig:renders_2d_gt_boxes}
   \vspace*{-4mm}
\end{figure*}

\textbf{Biased initialization:}
This simple initialization scheme improves convergence behavior and training performance without requiring a large dataset from which to learn a shape prior. In real-world outdoor scenes, most of the \stuff volume is empty space. By contrast, most of the volume inside each \things object bounding box is non-empty. We incorporate this prior directly by biasing the density prediction layer of the MLPs. For the \stuff MLP, we initialize our bias to $-5.0$, whereas for all \thing MLPs, we initialize the final bias layer to $0.1$. 
Furthermore, for all MLPs, we use the \textit{softplus} activation for the fully connected layer predicting the density outputs~\cite{Zheng15ijcnn_softplus}. We have found that this simple injection of prior knowledge via initializing the bias values is quite effective and robust compared to random initialization, since dense content in \stuff can suppress gradients from distant objects.

\textbf{Category specific learned initialization:}
If a large shape collection is available for certain object categories, we can further improve on the bias initialization scheme. In particular, we use meta-learning~\cite{finn2017maml} to capture category-specific shape and appearance priors. This process is illustrated in \cref{fig:fed_avg}. First, we meta-learn category-specific initial weights by pre-training on ShapeNet~\cite{chang2015shapenet}. Then, we use these weights to initialize the \thing MLPs in our model when training on a novel scene.

To meta-learn a category-specific shape prior, we use the \texttt{FedAvg}~\cite{mcmahan2017fedavg} algorithm. This algorithm is known to be equivalent~\cite{jiang2019improving} to REPTILE~\cite{nichol2018reptile} meta-learning, used before for NeRF initialization in \cite{Tancik21cvpr_meta}.  
To do one meta-step of \texttt{FedAvg}, we independently optimize a set of MLPs, each on a separate ShapeNet shape. We then average the model weights across all NeRF models, and start another meta-step. \cref{fig:fed_avg} visualizes the evolution of the learned initialization across several outer loops of \texttt{FedAvg}. In our experiments, we pre-train using the 2D rendered car images~\cite{Sitzmann19neurips_srn} of ShapeNet~\cite{chang2015shapenet} to obtain the learned initialization model. When reconstructing a full scene, we then initialize the MLPs for each car instance track $\track$ to this initialization.

\section{Evaluations}\label{sec:evaluation}

We performed a series of experiments to evaluate our model on multiple computer vision tasks, including view synthesis, reconstruction, 2D panoptic segmentation, 2D depth prediction, and scene editing. See the supplemental video for comprehensive visualizations of our results. All experiments used either the KITTI \cite{Geiger12cvpr_KITTI}, Virtual KITTI \cite{Gaidon16cvpr_virtual_kitti, Cabon20arxiv_virtual} or the recent KITTI-360~\cite{kitti360} datasets. These datasets involves difficult forward facing cameras in complex outdoor dynamic scenes. KITTI-360 is the first benchmark that evaluates both the task of synthesising color and appearance images from novel views. Our model outperforms every other method in that leaderboard for both tasks as shown in \cref{tab:kitti360}. Since scenes in KITTI-360 chosen for the tasks are all static, we also evaluate our model on dynamic scenes from the KITTI~\cite{Geiger12cvpr_KITTI} dataset. Rendered views from these dynamic scenes are shown in \cref{fig:renders_2d_predicted_boxes} and \cref{fig:renders_2d_gt_boxes}. Additional results are available in \cref{sec:additional_results}. Below we evaluate our model for each task in more detail.


\begin{table}
\centering
\resizebox{\linewidth}{!}{ 
\begin{tabular}{lcc}
\hline
Method                                                                  & Semantics mIoU & Appearance PSNR \\ \hline
NeRF~\cite{Mildenhall20eccv_nerf} + PSPNet~\cite{zhao2017pyramid}       & 53.01          & 21.18      \\
FVS~\cite{Riegler2020ECCV} +  PSPNet~\cite{zhao2017pyramid}             & 67.08          & 20.00      \\
PBNR~\cite{kopanas2021point} + PSPNet~\cite{zhao2017pyramid}            & 65.07          & 19.91      \\
Mip-NeRF~\cite{Barron21iccv_Mip_NeRF} + PSPNet~\cite{zhao2017pyramid}   & 51.15          & 21.54      \\
\textbf{Ours}                                                           & \textbf{74.28} & \textbf{21.91} \\ \hline
\end{tabular}
} 
\vspace*{-2mm}
\caption{
Results on novel view color and semantic synthesis tasks on KITTI-360~\cite{kitti360} dataset. Rendered semantic segmentation and color images from our model is the best performing method for both the tasks in KITTI-360 leaderboard.
} 
\label{tab:kitti360}
\end{table}
\begin{table}
\centering
\resizebox{\linewidth}{!}{ 
\begin{tabular}{lccccc}
\multicolumn{1}{l|}{}          & SRN~\cite{Sitzmann19neurips_srn}   & NeRF~\cite{Mildenhall20eccv_nerf}  & NeRF + time & NSG~\cite{nsg21cvpr}   & Ours           \\ \hline
\multicolumn{1}{l|}{PSNR $\uparrow$}      & 18.83 & 23.34 & 24.18       & 26.66 & \textbf{27.48} \\
\multicolumn{1}{l|}{SSIM $\uparrow$}      & 0.590 & 0.662 & 0.677       & 0.806 & \textbf{0.870}   \\
\end{tabular}
} 
\vspace*{-2mm}
\caption{Comparison of image reconstruction quality in dynamic KITTI scenes following the the experiement setup of NSG~\cite{nsg21cvpr}. 
} 
\label{tab:quantitative_kitti_nsg}
\vspace*{-4mm}
\end{table}

\textbf{Novel View Synthesis:}
How well a particular representation describes a scene is reflected in the quality of rendered views. As shown in~\cref{tab:kitti360}, color images rendered by our model achieves the best PSNR and is competitive with latest view synthesis models~\cite{Mildenhall20eccv_nerf,Riegler2020ECCV,kopanas2021point,Barron21iccv_Mip_NeRF}. Since the scenes used in KITTI-360~\cite{kitti360} novel view synthesis task are all static, we attribute the improved performance to our model benefiting from separate object-aware MLPs and incorporation of category-level priors. To study the novel view synthesis capabilities of our model on dynamics scenes, we also experiment on several dynamic scenes from KITTI dataset as shown in leftmost columns of \cref{fig:renders_2d_predicted_boxes} and \cref{fig:renders_2d_gt_boxes}. Notice that the rendered color images accurately captures the moving vehicles in the scene. A quantitative analysis of synthesized colored images for dynamic KITTI scenes is available in~\cref{tab:quantitative_kitti_nsg}, wherein we follow the same experimental setup as described in Sec.~5.2 of Ost~\etal~\cite{nsg21cvpr}. As expected, our method significantly outperforms representations like SRN~\cite{Sitzmann19neurips_srn} and NeRF~\cite{Mildenhall20eccv_nerf} which rely on static world assumption. Note that our method also outperforms NSG~\cite{nsg21cvpr} even though our instance specific object MLPs are much smaller ($10\times$ fewer FLOPs) compared to those used in NSG~\cite{nsg21cvpr}. The improvement over NSG also demonstrates the advantage of incorporating category-specific priors derived from meta-learning with images from ShapeNet.


\begin{table}
\centering
\resizebox{\linewidth}{!}{ 
\begin{tabular}{lcc}
\toprule
Method                                    & mIoU            & PQ\\ \midrule
2D Deeplab~\cite{Cheng20cvpr_Panoptic} on ground-truth RGB       & 49.9            & 43.2 \\
2D Deeplab~\cite{Cheng20cvpr_Panoptic} on NeRF rendered RGB      & 32.1            & 24.9 \\
\textbf{Ours} without \thing MLPs ($\approx$\textit{SemanticNeRF})                             & 45.3            & - \\
\textbf{Ours}                             & \textbf{56.5}   & \textbf{45.9} \\ \bottomrule
\end{tabular}
} 
\vspace*{-2mm}
\caption{
Ablation study of the rendered segmentation quality on KITTI. Our model achieves better segmentation in terms of mean IoU, panoptic quality (PQ)~\cite{kirillov2019panoptic}. Also notice that without \thing MLPs, the segmentation quality is significantly worse and misses all dynamic objects. This demonstrates the advantage of our model over non object-aware representations like \textit{SemanticNeRF}~\cite{Zhi21iccv_SemanticNeRF}.
} 
\label{tab:segmentation_2d}
\end{table}
\begin{figure}
  \centering
   \includegraphics[width=1.0\linewidth]{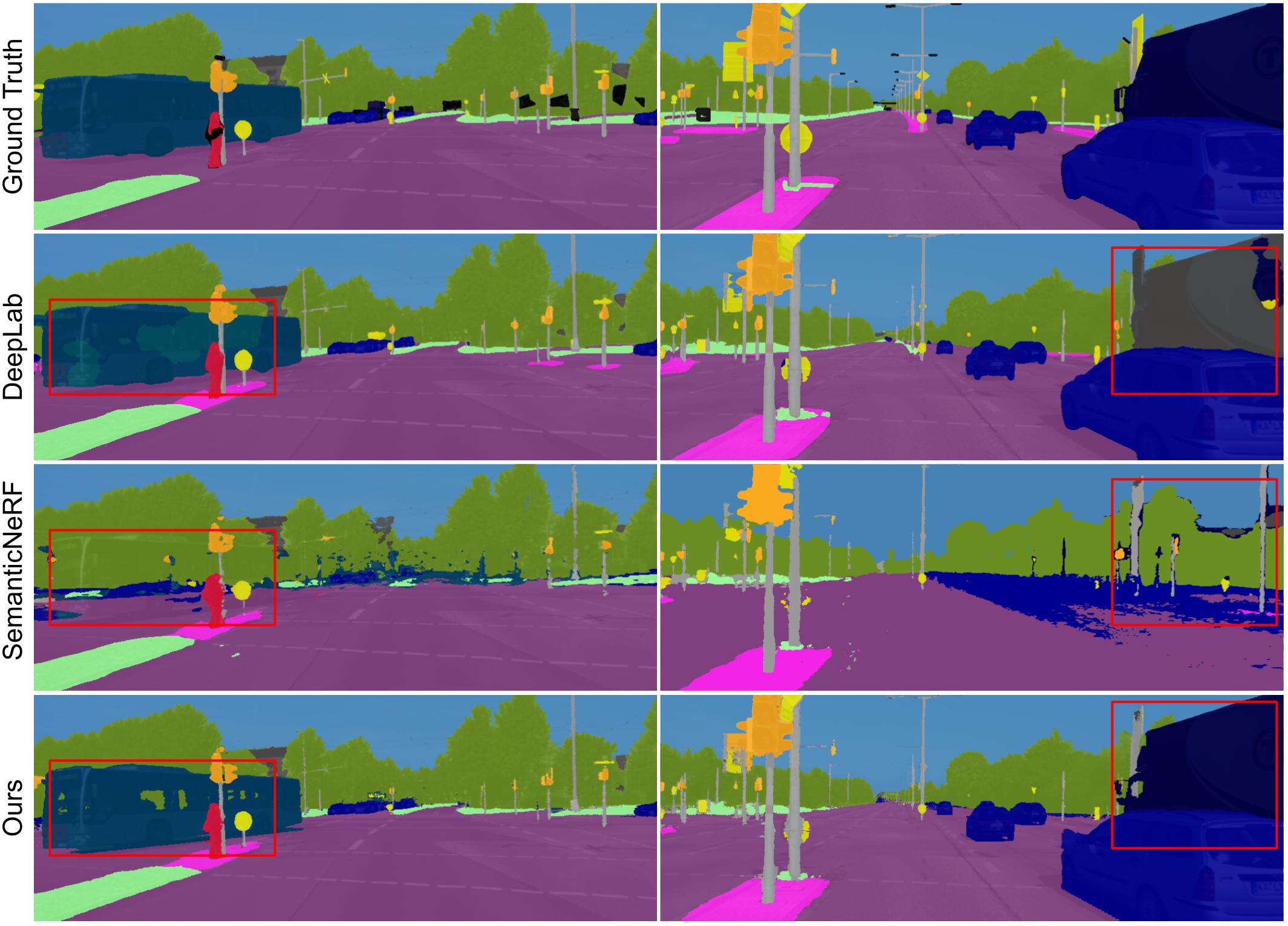}
   \caption{Comparison of semantic segmentation output from Panoptic-DeepLab~\cite{Cheng20cvpr_Panoptic} on ground-truth RGB image, SemanticNeRF~\cite{Zhi21iccv_SemanticNeRF}, and the semantic segmentation rendered from \textbf{our} model from the same view. Segmentation produced by \textbf{our} model is significantly better as highlighted by the red boxes in the figure.}
   \label{fig:ours_vs_deeplab}
   \vspace*{-5mm}
\end{figure}

\begin{figure*}
  \centering
   \includegraphics[width=1.0\linewidth]{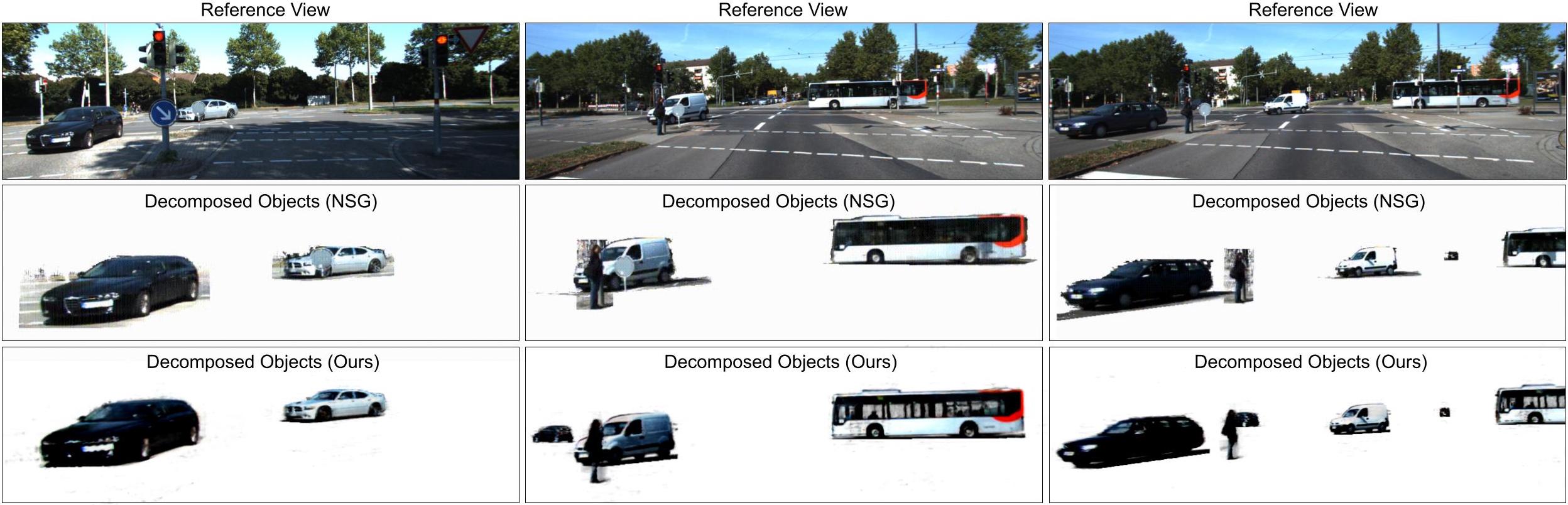}
   \vspace*{-6mm}
   \caption{Scene decomposition comparison.  Top row: reference views. Middle and Bottom: renderings of objects (without background stuff) of NSG~\cite{nsg21cvpr} and PNF (ours) model respectively. Note that the traffic sign-posts in front of the cars are entangled with the rendered cars in NSG, but not in our results. Also the windows of the bus are correctly reconstructed as translucent by our model.}
   \label{fig:scene_decomp_comp}
   \vspace*{-2mm}
\end{figure*}
\begin{figure}
    \includegraphics[width=1.0\linewidth]{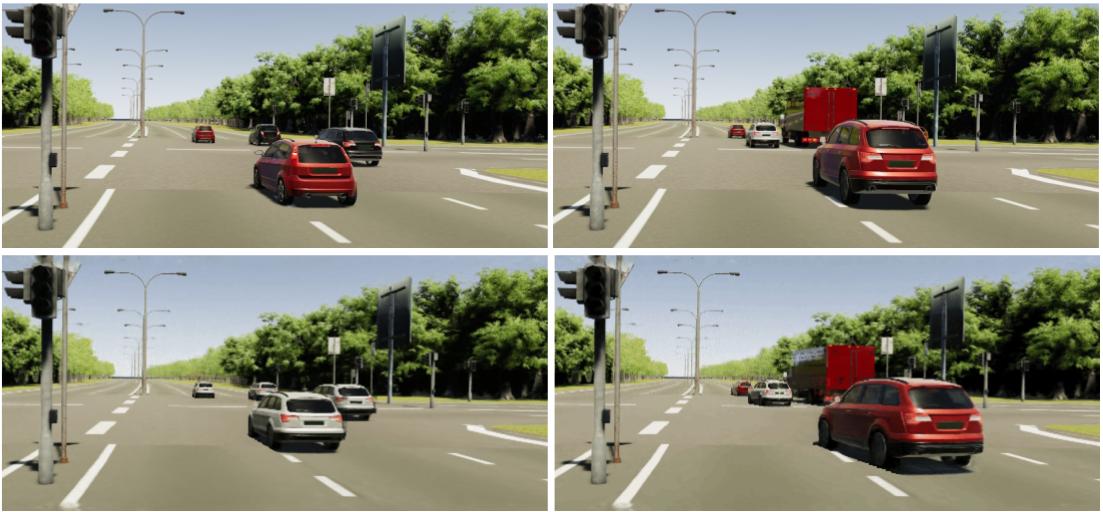}
   \vspace*{-6mm}
   \caption{Scene editing. We manipulate input ground truth images from KITTI Virtual~\cite{Gaidon16cvpr_virtual_kitti} (top) by cloning all cars to be the same (lower left) or changing their orientations (lower right).}
   \label{fig:scene_edit}
   \vspace*{-2mm}
\end{figure}

\textbf{Panoptic Segmentation:}
Semantic and instance segmentation of an arbitrary view can be obtained from our model by simple rendering (see \cref{eq:render}) along the desired view. The two right columns of \cref{fig:renders_2d_predicted_boxes} and \cref{fig:renders_2d_gt_boxes} demonstrates the rendered semantic and instance segmentation images from our model. As shown in \cref{tab:kitti360}, our model achieves \textit{state-of-the-art} 74.28 mIoU for the task of novel view semantic segmentation on KITTI-360. One approach of generating segmentation images for any arbitrary view is to first synthesize color image from a desired view using view synthesis methods like ~\cite{Mildenhall20eccv_nerf,Riegler2020ECCV,kopanas2021point,Barron21iccv_Mip_NeRF}; followed by 2D image segmentation~\cite{zhao2017pyramid,Cheng20cvpr_Panoptic}. However as demonstrated in \cref{tab:kitti360}, our unified model significantly outperforms all such two-stage baselines. Moreover the rendered segmentation images from our model are temporally consistent and works for dynamic scenes. We also perform a ablation study of the rendered segmentation images on dynamic scenes from KITTI. Quantitative and qualitative results are shown in \cref{tab:segmentation_2d} and \cref{fig:ours_vs_deeplab} respectively. Our model significantly out performs (+9.2 mIoU) non object-aware models like SemanticNeRF~\cite{Zhi21iccv_SemanticNeRF}, since they cannot model dynamic objects. Our model also improves upon single image state-of-the-art segmentation models~\cite{Cheng20cvpr_Panoptic} by fusing information from multiple views.

\textbf{2D Depth Estimation:} We also demonstrate rendered depth images from our model, obtained by over compositing point depths along rays with opacity values as described in~\cref{eq:render}. Rendered depth images are shown in the second columns of ~\cref{fig:renders_2d_predicted_boxes,fig:renders_2d_gt_boxes}, along with other outputs of the model. Please note the sharp reconstruction of shapes of moving cars on the road. Of course, those cars would be missing or blurred in standard NeRF models. This demonstrates the benefits of our object-aware approach that handles dynamic scenes and instance-specific object MLPs that accurately captures each object instance.

\textbf{Object Decomposition:}
\cref{fig:scene_decomp_comp} shows visualizations of how images from a dynamic KITTI scene are decomposed into MLPs representing different object instances. These images are rendered without the (stuff) background. Compared to NSG~\cite{nsg21cvpr}, our method does a better job of disentangling objects from the (stuff) background. In particular, note that the occluding traffic sign-posts in front of the car are entangled with the rendered cars in NSG, but not in our results. Better object decomposition from our model is also important for the scene editing task discussed below.

\textbf{Scene Editing:}
Since our model separates objects from the background and builds a full 3D radiance field for each object, it is possible to edit images using the model by removing objects, adding new objects, and transforming object bounding boxes and poses. \cref{fig:scene_edit} shows few scene editing examples on Virtual KITTI dataset. The top row shows original images, and the bottom row shows edits. In bottom-left, we demonstrate \textit{cloning} of cars by replicating the weights of all object MLPs to a same car. In bottom right of the figure we independently rotate each vehicle object.



\section{Limitations}
\label{sec:limitations}
Like most other NeRF-style methods, our model is compute-intensive and hence currently only suited for offline applications. However, we expect advances in neural rendering \cite{muller2022instant,tewari2021advances} will alleviate some of these speed issues in near future. It also does not incorporate more complex light transport effects, such as shadowing, under object motion.
Our framework optimizes and corrects bounding box poses from  noisy 3D object detection and tracking, but has not been designed to handle other errors such as missing and duplicate detections and incorrect class predictions. 
Finally, our framework does not handle deformable objects and is restricted to scenes with rigid moving objects.

\section{Conclusion}
\label{sec:conclusion}
This paper presents Panoptic Neural Fields (PNF), an object-aware neural scene representation that decomposes a scene into a set of MLPs associated with object instances (\things) and the background (\stuff). Our model learns a 4D panoptic radiance representation of dynamic scenes from images alone. This representation can be queried to obtain the color, density, instance, and category label of any 3D point over time. Several tasks like scene editing, view synthesis, panoptic segmentation are derived by simply rendering the representation from the desired views. Results of experiments on several KITTI scenes demonstrate state-of-the-art performance for novel view synthesis and panoptic segmentation for challenging outdoor scenes with multiple dynamic objects.
\pagebreak

\appendix

\begin{figure*}
  \centering
  \includegraphics[width=\linewidth]{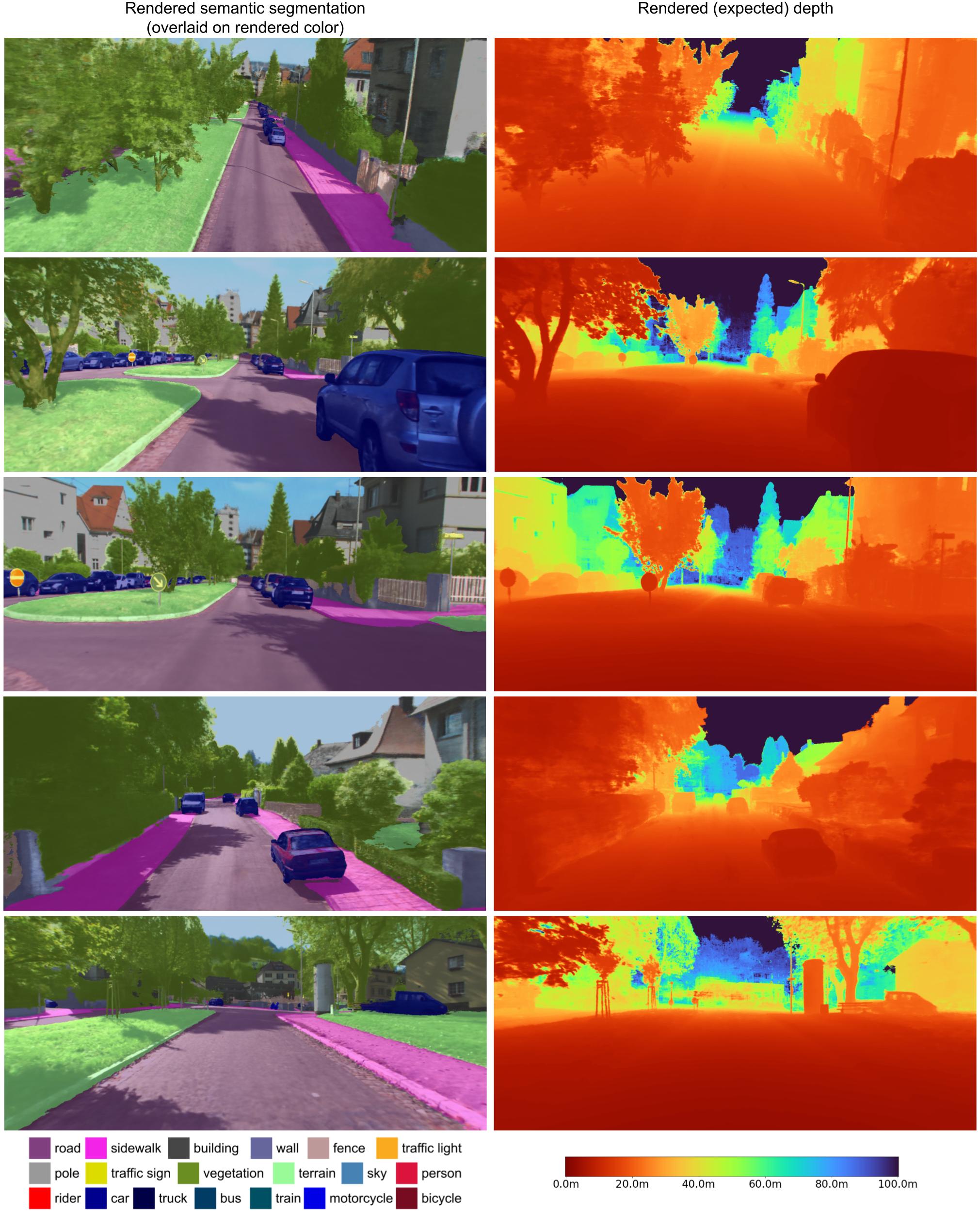}
   \caption{Rendered novel views (semantic segmentation and depth) from our Panoptic Neural Field models trained on KITTI-360~\cite{kitti360} scenes. Only the forward facing cameras (captured at $\approx$5Hz) are used for these results. Note that even thin structures like lamp poles and sign-posts are accurately reconstructed and segmented.}
   \label{fig:nvs_kitti360_results}
\end{figure*}
\begin{figure*}
  \centering
  \includegraphics[width=\linewidth]{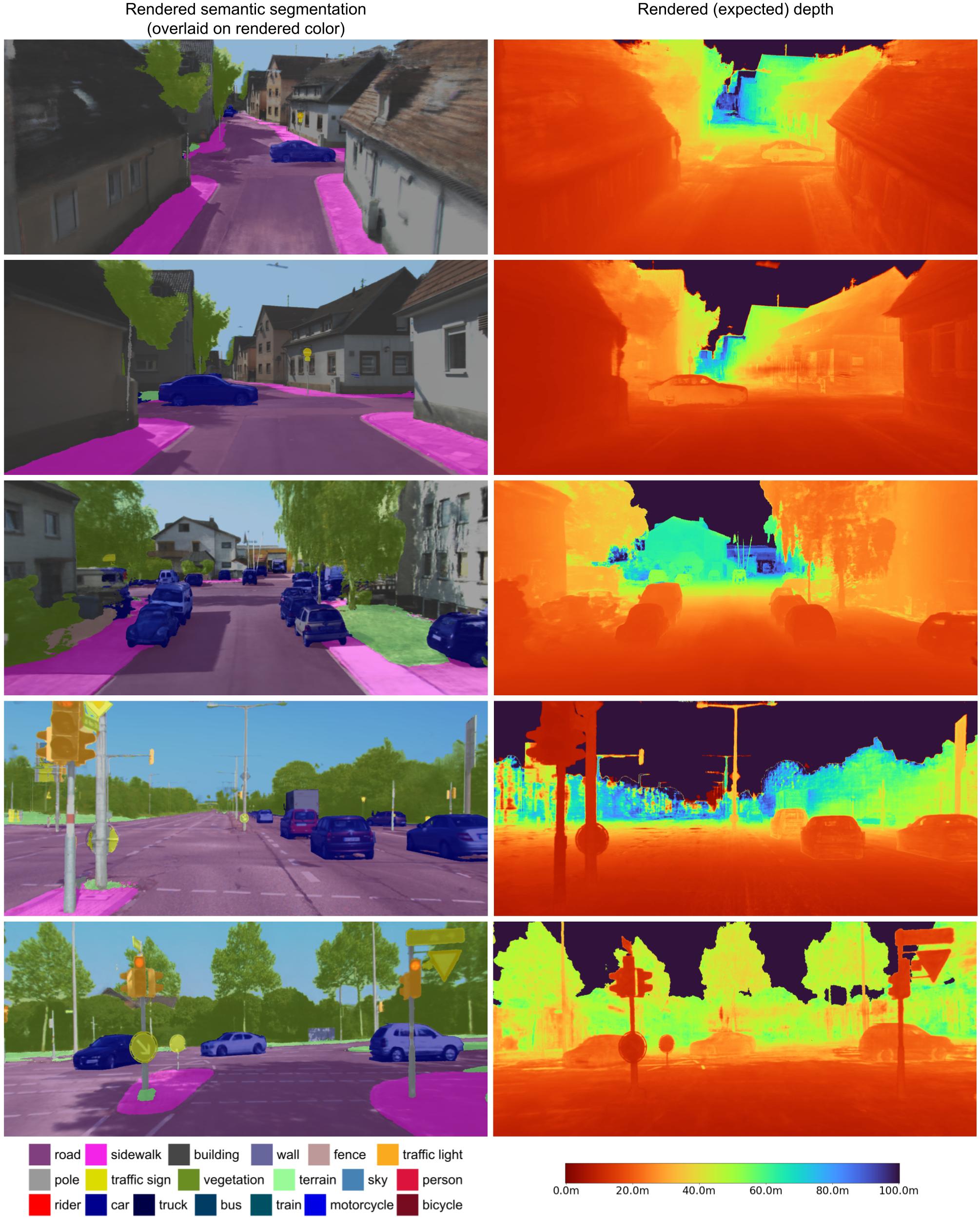}
   \caption{Rendered novel views (semantic segmentation and depth) from our Panoptic Neural Field models trained on KITTI~\cite{Geiger12cvpr_KITTI} scenes. Only the forward facing cameras (captured at $\approx$ 10Hz) are used for these results. Note that several of the scenes have moving cars but are still accurately reconstructed and segmented.}
   \label{fig:nvs_kitti_results}
\end{figure*}
\begin{figure*}
  \centering
  \includegraphics[width=\linewidth]{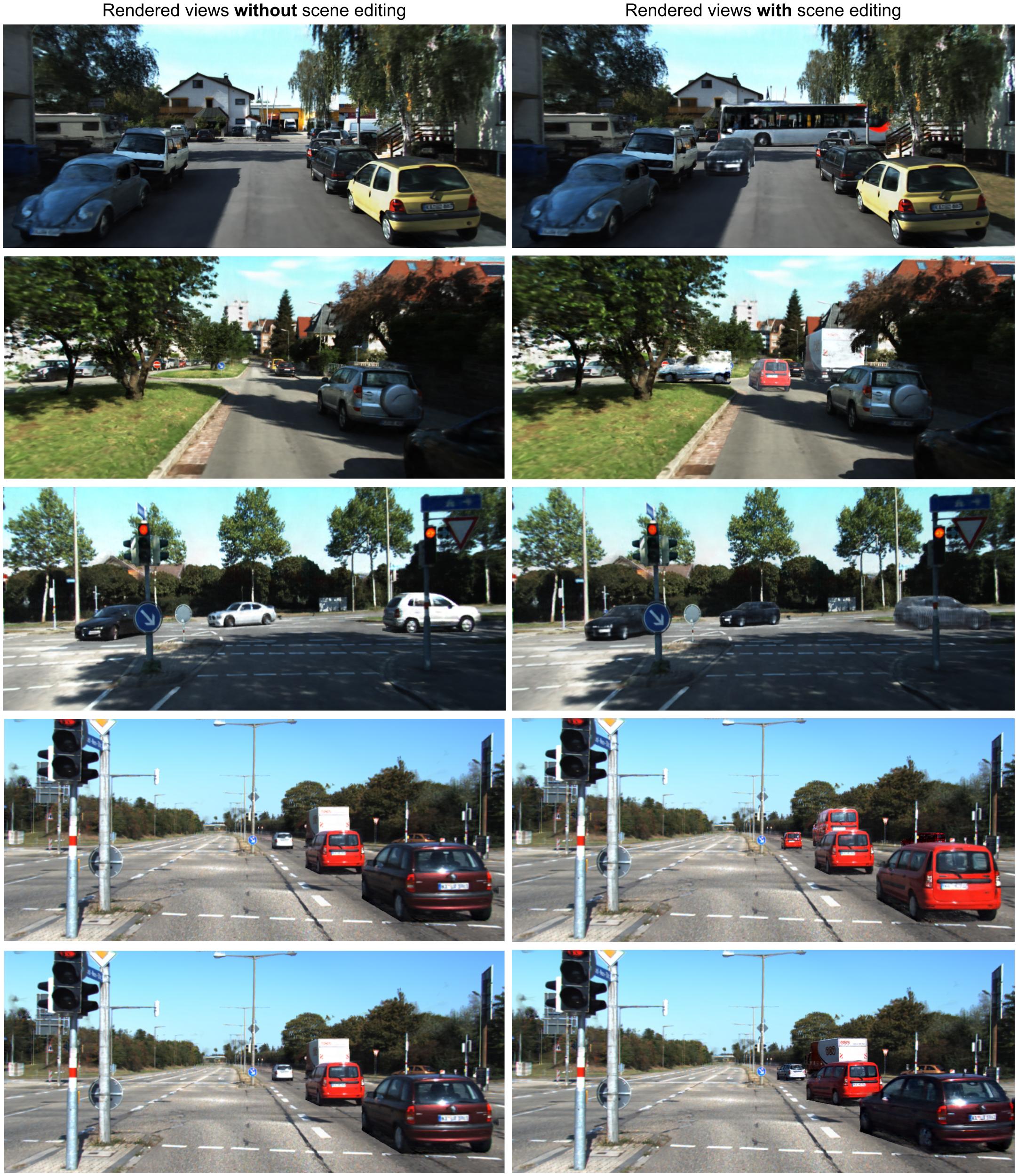}
   \caption{Scene editing results. The left column shows rendered color images along  novel viewpoints of our panoptic neural field representations learned on various scenes. Rendered color images from the same view but with edited scene representation are shown in right column. In the first two rows, new objects are introduced to the scene. The third and fourth row shows \textit{cloning} results wherein all \thing MLP weights are duplicated to that of a single object in the scene. The last row shows manipulation of 3D pose of the objects.}
   \label{fig:kitti_scene_editing}
\end{figure*}

\begin{figure*}[htpb]
\centering
  \includegraphics[width=1.0\linewidth]{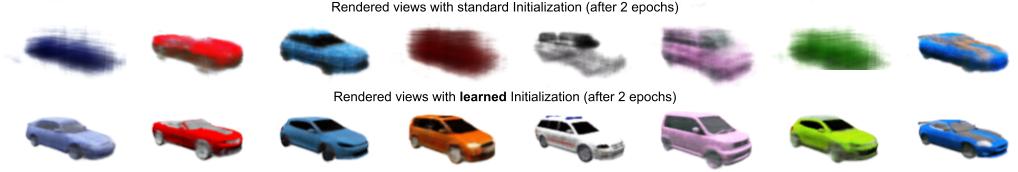}
\caption{Comparison of rendered views with different initialization methods on ShapeNet dataset. \textbf{Top row:} Standard xavier (glorot)~\cite{glorot2010understanding} initialization. \textbf{Bottom row:} Learned initialization via federated averaging. In both cases models were \textbf{only} optimized for two full epochs after the respective initialization. All models were trained on at-least 50 views. See \cref{fig:sparse_views_shapenet} for analysis with sparse views.}
\label{fig:convergence_shapenet}
\end{figure*}
\begin{figure*}[htpb]
\centering
  \includegraphics[width=1.0\linewidth]{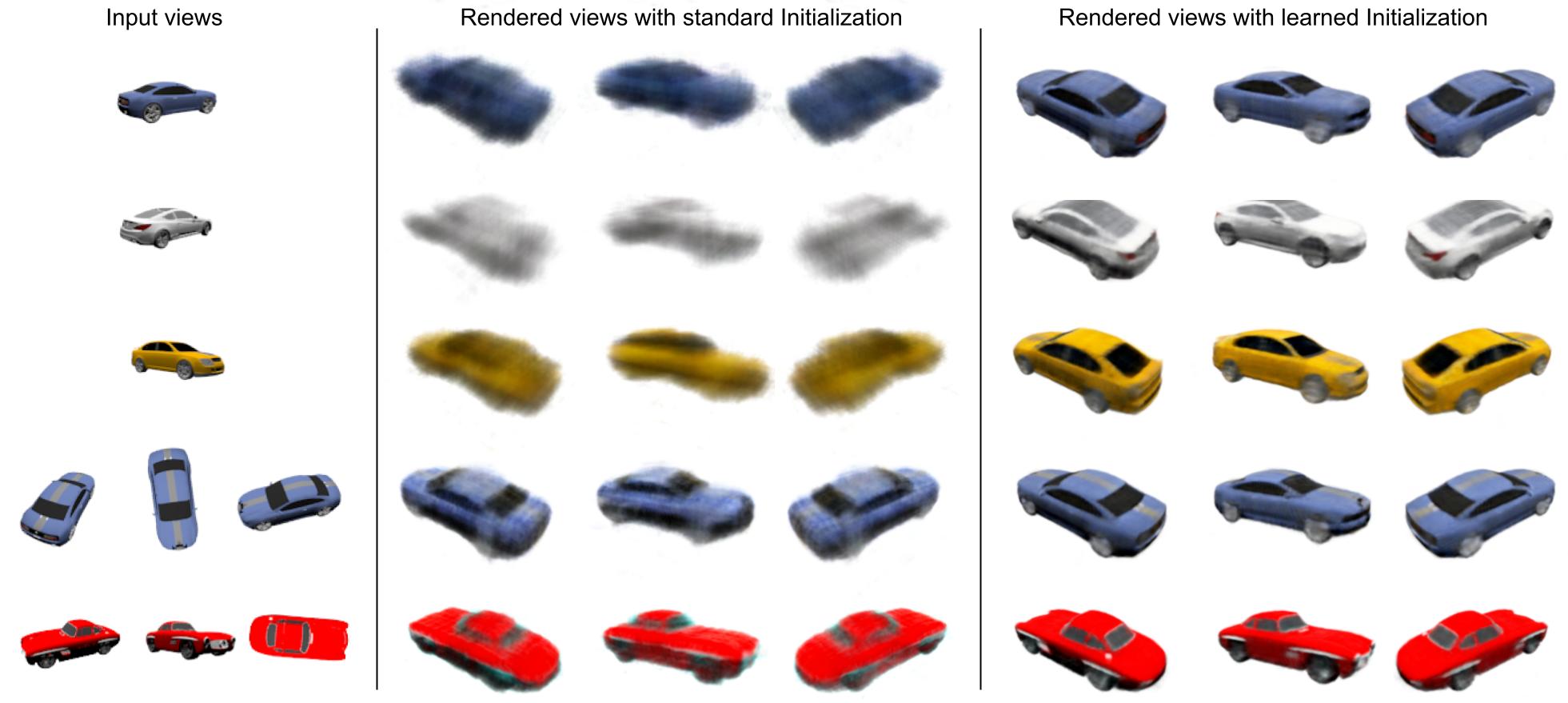}
\caption{Comparison of the effect of initialization methods on training \thing models on ShapeNet dataset with \textbf{very sparse} input views. We demonstrate using 1 or 3 image views as input to the model. The input views are shown left. Rendered views of the model when using the standard xavier (glorot)~\cite{glorot2010understanding} initialization are shown in the middle columns. Rendered views with our proposed learned initialization are shown in the right columns. Even with just one image input, our model can reconstruct the cars pretty well.}
\label{fig:sparse_views_shapenet}
\end{figure*}

\section{Additional Model Details}
In this section we provide additional details for training and inference of the proposed panoptic neural field model described in \cref{sec:method}.

\subsection{Network Architecture and Training Details}
For \stuff MLP, we use $8$ hidden layers of width $256$. For \thing MLP, we use MLP with $4$ hidden layers and width $128$. We use positional encoding of $L=10$ frequencies to encode position coordinates for \stuff MLP and $L=6$ frequencies to encode position coordinates in object coordinate space for each \thing MLP.

The density and semantic output does not depend on view directions, whereas the color output is additionally conditioned on view directions similar to \cite{Mildenhall20eccv_nerf}. This encodes the assumption that structure and semantics of the static background and individual objects only varies with position coordinates in their respective coordinate spaces. We use $L=2$ frequencies to encode the view directions before feeding them to the \stuff and \thing MLPs. For view directions, we find it beneficial to gradually activate the encoding frequencies over the course of the optimization, similar to~\cite{lin2021barf,Park21iccv_nerfies}.

To conserve memory we do not perform hierarchical sampling~\cite{Mildenhall20eccv_nerf}. Instead we sample additional points with stratified sampling strategy along rays while training and inference. So unlike  ~\cite{Mildenhall20eccv_nerf} which uses a pair of MLPs corresponding to coarse and fine sampling, only one set of MLPs are required in our model. We used 1024 samples per ray in our experiments on KITTI dataset. 

As described in \cref{sec:model_train}, we use both color loss and semantic loss. Since semantic loss is realized with softmax cross-entropy function compared to mean squared error function to realize the RGB (color) loss, we scale the semantic loss lower to ensure that the semantic loss does not dominate the training process. Apart from the color loss and semantic loss, the optimization objective also enforces the constraint that rotation component $\rotation \in \real^{3 \times 3}$ of the optimized tracks are valid $\mathrm{SO}(3)$ rotation matrices using the SVD orthogonalization~\cite{levinson2020analysis} method. We use Adam as our optimizer for all our training routines including optimizing Panoptic Neural Field model to a new scene and also during inner loops of \texttt{FedAvg} based meta-learning (see \cref{sec:model_init}) for learned initialization.

\subsection{Model Initialization Details}
As described in \cref{sec:model_init}, we incorporate priors via initialization. For \textit{cars} and \textit{vans}, we use category specific initialization of \thing MLP weights. This category specific initialization is meta-learned from 2D rendered images of 3D \textit{cars} models from ShapeNet dataset. For all other object categories, we use the biased initialization described in \cref{sec:model_init}. We use a simplified federated averaging algorithm~\cite{mcmahan2017fedavg} as described in \cref{alg:fedavg} to realize our category specific learned initialization. Also see \cref{fig:fed_avg} which visualizes the evolution of the learned initialization across several outer loops of \texttt{FedAvg}. This algorithm is known to be equivalent~\cite{jiang2019improving} to REPTILE~\cite{nichol2018reptile} meta-learning. The main advantage of the simple \texttt{FedAvg} is that it allows decentralized federated training. In \cref{fig:fed_avg}, we assumed simple SGD as the optimization algorithm, but can be easily adapted for other optimizers.

\begin{algorithm}
\caption{\texttt{FedAvg}. The $K$ clients each working on images of an unique object instance are indexed by $k$; $\mathcal{D}_k$ is the data (bundle of rays  corresponding to every observed pixel) for object instance $k$; $B$ is the local minibatch size \ie number of rays used per batch for inner epochs on each client; $E$ is the number of inner (local) epochs, and $\eta$ is the learning rate; the network parameters (MLP weights) is denoted by $\theta_t$ after $t$ outer epochs on the server.}\label{alg:fedavg}
\begin{algorithmic}
\Procedure{ServerUpdate}{}
\State initialize $\theta_0$
\For{each outer epoch $t = 1, 2, \dots$}
    \For{each client $k \in 1, 2, \dots, K$ \textbf{in parallel}}
        \State $\theta_{t+1}^{k} \leftarrow \textproc{ClientUpdate}(k, \theta_t)$
    \EndFor
    \State $\theta_{t+1} \leftarrow \frac{1}{K} \sum_{k=1}^{K} \theta_{t+1}^{k}$
\EndFor
\EndProcedure
\\
\Function{ClientUpdate}{$k,\theta$}
\State $\mathcal{B} \leftarrow (\text{split $\mathcal{D}_k$ into batches of ray bundles of size \textit{B}}$)
\For{each inner epoch $i$ from 1 to $E$}
    \For{batch $b \in \mathcal{B}$}
        \State $\theta \leftarrow \theta - \eta \triangledown \ell (\theta;b)$
    \EndFor
\EndFor
\State return $\theta$ to server
\EndFunction
\end{algorithmic}
\end{algorithm}

\section{Additional Results}\label{sec:additional_results}

In this section we provide additional results on KITTI \cite{Geiger12cvpr_KITTI} and KITTI-360~\cite{kitti360} for the tasks of novel view synthesis of color, semantics and depth images, along with scene editing. We also evaluate the benefits of our proposed category specific meta-learned initialization of \thing MLP weights on the ShapeNet~\cite{chang2015shapenet} dataset.

\subsection{Novel view renderings}
Additional qualitative results of our model on KITTI \cite{Geiger12cvpr_KITTI} and KITTI-360~\cite{kitti360} are shown on \cref{fig:nvs_kitti_results} and \cref{fig:nvs_kitti360_results} respectively. Just like our experiments in \cref{sec:evaluation}, we only used the forward facing cameras images for these results. Note that even though KITTI-360~\cite{kitti360} dataset provides side facing fisheye camera images, only the forward facing stereo images are available for the novel view test sequences used for the experiments. Each row of \cref{fig:nvs_kitti_results} and \cref{fig:nvs_kitti360_results} shows a novel view of a scene generated by rendering the panoptic neural field representation of that scene. The left columns shows the rendered semantic segmentation overlaid on top of the rendered color image so that it is easier to judge the segmentation quality. The colormap used for visualizing the segmentation and depth images are shown at the bottom. The corresponding rendered depth images from those views are shown in the right columns. Note that even the difficult thin structures like lamp poles and sign posts are both accurately reconstructed and segmented by our model. Also note the accurate reconstruction and segmentation of multiple moving cars in \cref{fig:nvs_kitti_results}. Additionally, rendered color images from some novel viewpoints are also available the left column of \cref{fig:kitti_scene_editing}.

\subsection{Scene Editing}\label{sec:additional_scene_editing}
Since our proposed panoptic neural field scene representation is object aware, it allows seamless manipulation and editing of different objects present in the scene. In addition to the scene editing results on Virtual KITTI~\cite{Cabon20arxiv_virtual} dataset discussed in \cref{sec:evaluation} and \cref{fig:scene_edit}, we show scene editing results on KITTI and KITTI-360 datasets in \cref{fig:kitti_scene_editing}. Each row in \cref{fig:kitti_scene_editing} shows rendered color images along some novel view of both the original (left) and edited (right) scene representations. More specifically we show adding and removing new objects into the scene, changing the 3D pose of objects, and object \textit{cloning} where the \thing MLP parameters are replicated for all objects in the scene.

\subsection{Benefits of our learned initialization}

In real world scenes, objects are often captured from a sparse set of views. For example in self driving car scenes like KITTI, most objects (\eg cars) often get a limited set of views from one side only. Thus incorporating prior knowledge becomes important for completeness and accurate reconstruction.

We learn the category specific priors from a large collection of objects on ShapeNet, as part of a separate meta-learning process and distill that knowledge as initialization when training on a novel scene. Thus our inference time scene representation network is more efficient and only focus on the individual set of object instances present in the scene. As demonstrated in \cref{tab:quantitative_kitti_nsg}, our model does a better job in reconstructing images of a dynamic scene compared to other object aware approaches like NSG~\cite{nsg21cvpr}, even though we use a much smaller MLP (10x fewer FLOPs) per object.

The learned initialization also provides other benefits like faster convergence and better completeness when reconstructed from sparse partial observations. We demonstrate these two benefits on ShapeNet~\cite{chang2015shapenet} dataset in \cref{fig:convergence_shapenet} and \cref{fig:sparse_views_shapenet}. Specifically, we used rendered images of \textit{cars} from ShapeNet~\cite{chang2015shapenet} provided by~\cite{Sitzmann19neurips_srn}. 

\cref{fig:convergence_shapenet} qualitatively compares the rendered color images when using learned initialization over the standard xavier (glorot)~\cite{glorot2010understanding} initialization after two full epochs of training. For this experiment each model has at-least 50 input views. As seen in \cref{fig:convergence_shapenet}, the proposed initialization offers clear benefits in terms of faster convergence even when there is a dense set of input views.

The advantage of the learned initialization over standard initialization is more pronounced when we have few sparse input views of an object. This is demonstrated in \cref{fig:sparse_views_shapenet}. Using the proposed learned initialization, our model can reconstruct novel object instances even with just a single image as input. As shown in \cref{fig:sparse_views_shapenet}, even when only a partial view of the objects are used as input, the category specific object priors distilled via the initialization results in a more complete reconstruction.  

\subsection{Video Results}
We also encourage the readers to also look at supplemental videos demonstrating results of our framework and a overview of the method. Most of our results on dynamic scenes are better visualized in the video.

\section{Potential Negative Societal Impact}

Our contribution is an intermediate representation for comprehensive 3D scene understanding. We believe this can enable applications with a beneficial impact on society. However, it could also enable applications with potential negative impact. While it is impossible to anticipate all possible such applications, we discuss a few below.

Because our method supports comprehensive tracking of objects and people, it could be extended for use in crowd monitoring, traffic density reports and beneficial applications stemming from that. However, it could also be incorporated into surveillance systems.  We will include stipulations in the license agreement for the code limiting its applications to academic research.

In addition, because our methods support view synthesis of 3D scenes, it is conceivable that it could be used to create imagery of fictional events, with the potential to disseminate fake news and/or propaganda.  Because our method supports scene editing, actual events could be altered and used in similar ways.  Of course, we will clearly mark all images generated by our system as ``synthetic.''  Additionally, we will include a requirement to do the same in the download instructions for our code.

Mitigation of the above issues is hard: many computer vision contributions are intermediate representations like ours. Segmentation, feature tracking, and object recognition can be put together into diverse functioning applications. As a profession, we should strive for the ethical application of these new technologies.

{
    \small
    \bibliographystyle{ieee_fullname}
    \bibliography{main}
}

\end{document}